\documentclass{article}
\usepackage{spconf,amsmath,graphicx}
\usepackage[font=small,labelfont=bf]{caption}
\usepackage[caption=false]{subfig}
\usepackage[colorlinks,citecolor=red,urlcolor=blue,bookmarks=false,hypertexnames=true, pagebackref=true]{hyperref}            
\usepackage{makecell}
\usepackage{float}

\title{Tackling electrode shift in gesture recognition with HD-EMG electrode subsets}
\name{Joao Pereira$^{1}$, Dimitrios Chalatsis$^{1}$, Balint Hodossy$^{1}$ and Dario Farina$^{1}$ \thanks{This project was supported by UK Research and Innovation [UKRI Centre for Doctoral Training in AI for Healthcare grant number EP/S023283/1] and Imperial-META Wearable Neural Interfaces Research Centre.
}}
\address{$^{1}$Imperial College London, Neuromechanics \& Rehabilitation Technology Group, UK }
\usepackage[font=small,labelfont=bf]{caption}

\begin{document}

%
\maketitle
\begin{abstract}
sEMG pattern recognition algorithms have been explored extensively in decoding movement intent, yet are known to be vulnerable to changing recording conditions, exhibiting significant drops in performance across subjects, and even across sessions. Multi-channel surface EMG, also referred to as high-density sEMG (HD-sEMG) systems, have been used to improve performance with the information collected through the use of additional electrodes. However, a lack of robustness is ever present due to limited datasets and the difficulties in addressing sources of variability, such as electrode placement. In this study, we propose training on a collection of input channel subsets and augmenting our training distribution with data from different electrode locations, simultaneously targeting electrode shift and reducing input dimensionality. Our method increases robustness against electrode shift and results in significantly higher intersession performance across subjects and classification algorithms.

\end{abstract}
{
\footnotesize
\hfill \hfill 

© 2024 IEEE.  Personal use of this material is permitted.  Permission from IEEE must be obtained for all other uses, in any current or future media, including reprinting/republishing this material for advertising or promotional purposes, creating new collective works, for resale or redistribution to servers or lists, or reuse of any copyrighted component of this work in other works.
}
\section{Introduction}
\label{sec:intro}

Surface electromyography (sEMG) measures muscular activity from the skin-surface. Thus, it has been extensively explored as a non-invasive method to decode movement intention for interfacing with virtual environments \cite{AR/VR}, neural prostheses \cite{prostheses} and exoskeletons \cite{exoskeletons}. Traditionally, the EMG control problem is modelled as gesture classification or joint-angle regression based on short sEMG intervals \cite{review}. Despite its prowess in laboratory conditions, pattern recognition sEMG-based control has not yet been widely adopted commercially due to limited robustness to sources of interference encountered in day-to-day use.

sEMG signals have multiple sources of variability, such as anatomical differences among subjects, non-stationarities due to force variations and limb position \cite{confounding}, as well as electrode placement. This prevents not only generalization across subjects, but also across sessions for a single subject without repeated calibrations \cite{capgmyo}. In particular, electrode shift, which refers to displacement in electrode position across recording sessions, results in significant drops in performance for even small distances. A 1-cm shift in a simple task with 10 gestures can result in an average accuracy drop of 25\% \cite{electrode-shift}. Interestingly, the authors also showed that this drop in performance is significantly reduced by introducing electrode-shifted data in the training distribution.

While deep learning has shown remarkable prowess in learning invariances to such perturbations, it relies on the existence of enough data to cover the input distribution with fidelity. Unlike other domains, such as computer vision or natural language processing, where deep learning has led to significant advancements, sEMG data collection requires time-consuming sessions, expensive equipment and trained personnel, making available data scarce. In practice, either simpler models are investigated with adaptive training \cite{capgmyo}, few-shot learning \cite{metalearning}, or traditional feature extraction pipelines are employed \cite{TD, ETD, ninapro, SampEn}.

Deep learning has also been used to leverage high-density sEMG (HD-sEMG) grids, which provide sEMG at higher dimensionality and spatial resolution \cite{capgmyo, metalearning}. However, despite the useful information, the data variability is even higher due to its high-dimensionality. Other methods to mitigate the effects of electrode shift have been explored, such as adopting a more robust electrode configuration \cite{interelectrode}, borrowing from CNNs in computer vision to learn HD-sEMG shift invariance \cite{translation-invariance}, as well as including electrode-shifted signals in the training distribution \cite{tackling-electrode-shift}. Training on multiple HD-sEMG channel subsets has been explored as a performance comparison to position verification \cite{PV}, and treating different HD-sEMG grids as a set of simultaneously acquired low-dimensional inputs has been explored in gait cycle prediction \cite{balint-annika}. 

Here, we propose training on subsets of HD-sEMG channels as an end-to-end solution for gesture recognition and evaluate its effect on intersession performance. The concept is to enable translation from robust algorithms trained on HD-sEMG datasets to simpler, low-density EMG systems, such as set-ups used in \cite{ninapro}. The findings suggest that using the remaining data for data augmentation increases the number of samples per trainable parameter, with a resulting gesture classifier that is more resistant to electrode shift. In summary:
\begin{enumerate}
    \item We show that by including subsets of channels at different placements during training (i.e. electrode shift data augmentation), classification algorithms become less sensitive to electrode shift, congruent with past literature \cite{electrode-shift, tackling-electrode-shift}.
    \item This augmentation results in higher intersession accuracy than training on all available channels at a lower computational complexity. 
\end{enumerate}
These results were demonstrated using the publicly available Capgmyo dataset \cite{capgmyo}. 

\section{METHODS}
\label{sec:meth}

\subsection{Dataset}
In this study, experiments are carried out using the raw HD-sEMG data from the Capgmyo dataset \cite{capgmyo}, which recorded signals via a differential, silver, wet electrode array as shown in Figure \ref{fig:subsets}, resulting in 128 simultaneous channel recordings acquired at 1000 Hz \cite{capgmyo-acquisition}. The dataset was split into 3 subsets. Experiment 1 used data from subset DB$_a$, containing single sessions from 18 subjects, whereas Experiment 2 used data from DB$_b$, containing two sessions for each of the 10 subjects. Due to data corruption in the last subject's recordings, this experiment was carried forward with the first 9 subjects. 

Each session consisted of 10 repetitions of 8 isotonic and isometric gestures ($G=8$): \textbf{(i)} raising thumb upwards, \textbf{(ii)} middle/index finger extension with flexion of remaining fingers, \textbf{(iii)} flexion of ring and little finger with extension of remaining fingers, \textbf{(iv)} thumb opposing base of little finger, \textbf{(v)} abduction of all fingers, \textbf{(vi)} finger flexion in a fist, \textbf{(vii)} pointing index and \textbf{(viii)} adduction of extended fingers. Each gesture was held for 3-10s.

Due to the reaction time of the subjects after receiving instructions, acquired labels may not perfectly match gestures. Therefore, as in \cite{capgmyo}, only the central 1s interval of each gesture was considered in this study. Time-window and stride durations of 256ms ($T=256$) and 15ms were used respectively. For Experiment 1, windows from half of the repetitions of each gesture were assigned to the training set and the remaining half to the test set, resulting in a train/test size of 1960 instances. For Experiment 2, the train/test sets were each of the two sessions of a subject, resulting in train/test sizes of 3920 instances.

\subsection{Electrode Subset Sampling}

\begin{figure}[]
  \includegraphics[clip,width=\columnwidth]{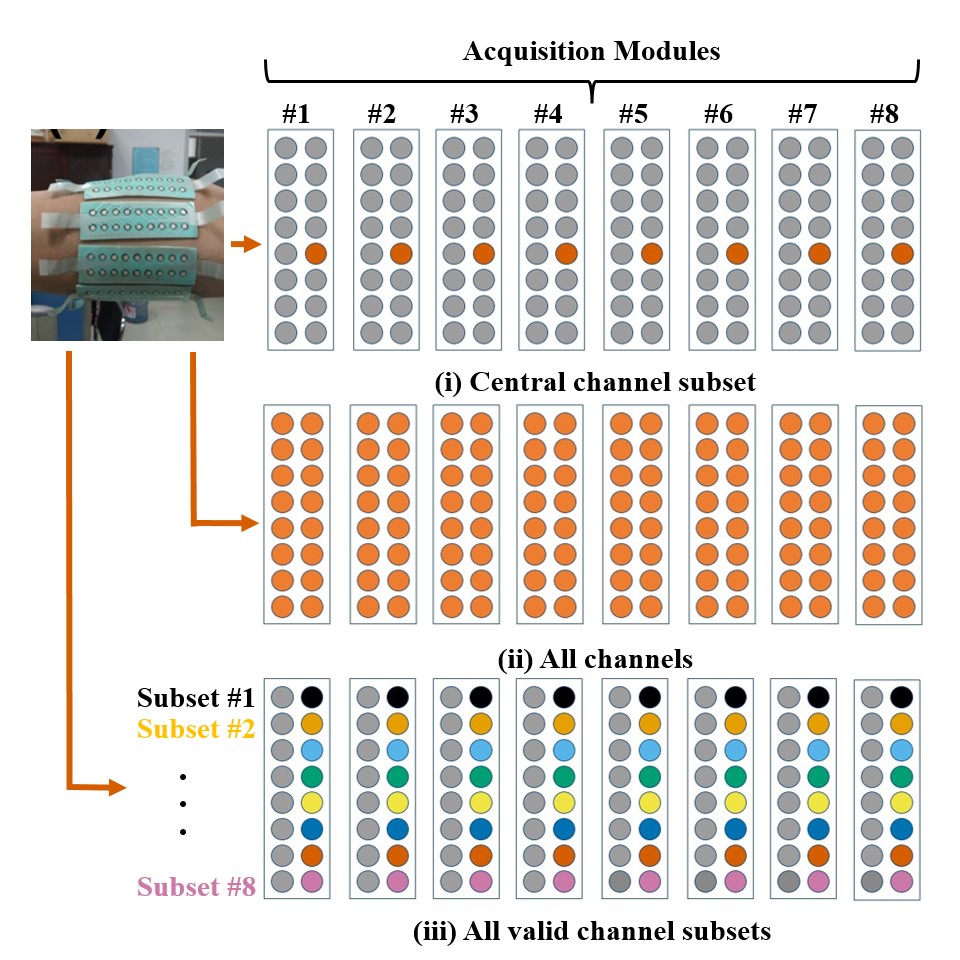}%
\caption{Schematic of channel subset sampling procedure applied to the Capgmyo dataset set-up. Three training conditions are considered: \textbf{(i)} training only on the central channel subset, where one channel per acquisition channel is considered; \textbf{(ii)} training/testing on all (128) channels; \textbf{(iii)} training on all valid channel subsets, where only proximal-distal translations are considered.}
\label{fig:subsets}
\end{figure}

Gesture classification is a supervised learning task with dataset $D = \{ (\mathbf{x}_{n}, \mathbf{y}_{n}) \}_{n=1}^{N}$ consisting of $N$ EMG time-windows $\mathbf{x} \in R^{Ch \times T}$ and gestures $\mathbf{y} \in R^{G}$, where the aim is to learn a mapping $f : R^{Ch \times T} \rightarrow R^{G}$. Given 128 bipolar channels, an electrode subset comprised eight channels $(Ch=8)$ evenly spaced along the circumference of the forearm. This mimicked set-ups such as in \cite{ninapro}. Given the grid dimensions of 8x16, there were eight possible channel subsets that could be selected when considering proximal-distal shifts, with the channel subset referred to as the \textit{central channel subset} (Figure \ref{fig:subsets}). Given the grid geometry, this provided a maximal shift of 2.4cm distally and 3.2cm proximally \cite{capgmyo-acquisition}. Clockwise/anticlockwise shifts along the arm circumference  were not included in this study. By treating electrode shift as a transformation $T\{.\}$ to the inputs, our method can be interpreted as a data augmentation technique for increasing the training set size and learning an electrode shift invariant mapping $ f(T \{ \mathbf{x} \})=f( \mathbf{x})$ for all valid EMG time-windows. 

\subsection{Preprocessing \& Machine Learning}

Power-line interference was reduced using a digital band-stop filter (45-55H Hz, second-order Butterworth) and all channels were standardized across gestures. As evident from the literature \cite{du2017surface,fan_csac-net_2022}, conventional deep learning approaches exhibit limited generalization capabilities within intersession settings. This is usually mitigated with techniques like transfer learning and meta learning \cite{fan_csac-net_2022}. Our primary focus was to enhance the performance of basic machine learning models for intersession scenarios, without relying on transfer learning methods.

\begin{figure}[t!]

\centering
\includegraphics[width=0.4\textwidth]{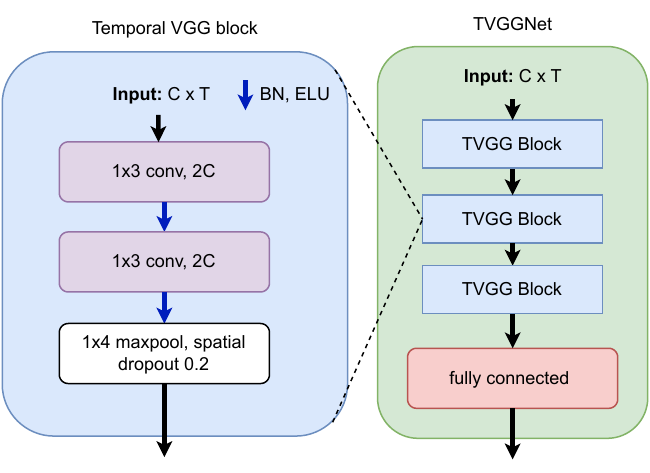}
\caption{Architecture of our temporal VGGNet. The network consists of a number of TVGG blocks consisting of two temporal convolutions, a maxpool and spatial dropout. Each TVGG block doubles the number of channels, while keeping the temporal dimension constant. }
\label{fig:TVGG}
\end{figure}

Thus, a network inspired by Visual Geometry Group (VGG) based architectures was implemented, with a detailed description provided in Figure \ref{fig:TVGG}. Our primary objective was assessing the effectiveness of our proposed training method across various machine learning tools rather than pursuing state-of-the-art results. To achieve this broad applicability, we opted for this foundational CNN architecture, upon which many more complex models have been built. This decision enhances the generalizability of our findings, primarily owing to the fundamental nature of the selected architecture, instead of them being  tied exclusively to a specific architectural choice. Furthermore, it is important for our experiments that our chosen architecture can adeptly handle varying input sizes, which is something this versatile architecture can perform well.

Four traditional state-of-the-art feature sets were also implemented: Hudgin's time-domain features (TD) \cite{TD}, the enhanced TD features (ETD) from  \cite{ETD}, the best performing features from the original NinaPro dataset publication \cite{ninapro}, and the SampEn feature set, which was shown to be the feature set with most stable performance across days \cite{SampEn}. Classification was carried out via linear discriminant analysis (LDA).

\section{EXPERIMENTAL ANALYSIS}
\label{sec:exp}

\subsection{Electrode Shift Generalization}
The first line of enquiry was to investigate whether the considered feature sets and network architectures can generalize to proximal-distal displacements with the inclusion of such displacements during training. Training and testing sets were created with a 50:50 split by including every other gesture repetition in each respective set. 

For all feature sets (and TVGGNet), four conditions were considered, where methods could be trained and/or tested on either data from the central subset (CS) or from all valid subsets (AVS):
\begin{enumerate}
    \item \textbf{(CS-CS)}: training/testing on the central subset for baseline performance
    \item \textbf{(AVS-AVS)}: training/testing on all valid subsets to validate whether models are expressive enough to capture full augmented distribution
    \item \textbf{(AVS-CS)}: training on all valid subsets and testing on central subset to evaluate whether performance drops at specific locations
    \item \textbf{(CS-AVS)}: training on central subset and testing on all valid subsets to evaluate the effect of electrode shift on performance in a usual experimental set-up
\end{enumerate}

The test performance averaged across subjects for each condition is shown in Table \ref{table:intra}. A one-way ANOVA test indicated that the implemented feature sets and the TVGGNet do not exhibit significantly different test accuracies ($F(3,284)=1.640, p=0.164$). Thus, the average score across feature sets and TVGGNet was considered in this analysis. Given that a Levene test did not support the assumption of homogeneity of variances across experiment conditions ($F(3, 68)=5.654, p =0.002$), non-parametric tests were used to compare test accuracies across experimental conditions. Wilcoxon signed-rank tests exhibited significant differences when comparing experimental conditions \textbf{(CS-CS)} and \textbf{(AVS-AVS)} ($V=171, p<0.001$), \textbf{(CS-CS)} and \textbf{(AVS-CS)} ($V=156, p<0.001$) and \textbf{(CS-CS)} and \textbf{(CS-AVS)} ($V=171, p<0.001$). However, it is important to note that while average test performance only reduces relative to \textbf{(CS-CS)} by 4\% and 2\% for \textbf{(AVS-AVS)} and \textbf{(AVS-CS)} respectively, there is a 25\% drop for \textbf{(CS-AVS)}. This suggests that training on all valid subsets has a regularizing effect across models, where marginal drops in performance are seen in controlled conditions and far larger improvements are seen in electrode-shift conditions. 

\begin{table}[t!]
\renewcommand{\arraystretch}{1.7}
\caption{Intrasession test accuracy across Experiment 1 settings and feature sets (and TVGGNet), averaged across subjects in DB$_{a}$.}
\footnotesize
\setlength{\tabcolsep}{3pt}
\begin{tabular}{ccccc}
\Xhline{1pt}
                        & CS-CS                 & AVS-AVS                 & AVS-CS                 & CS-AVS                 \\ \hline
TD                      & 0.920 $\pm$0.050  & 0.867 $\pm$ 0.047 & 0.896 $\pm$ 0.058 & 0.659 $\pm$ 0.104 \\ \hline
ETD                     & 0.944 $\pm$ 0.036 & 0.920 $\pm$ 0.033 & 0.930 $\pm$ 0.045 & 0.672 $\pm$ 0.104 \\ \hline
NinaPro                 & 0.909 $\pm$ 0.057 & 0.869 $\pm$ 0.050 & 0.891 $\pm$ 0.061 & 0.661 $\pm$ 0.098 \\ \hline
SampEn                  & 0.950 $\pm$ 0.043 & 0.904 $\pm$ 0.046 & 0.915 $\pm$ 0.053 & 0.646 $\pm$ 0.133 \\ \hline
TVGGNet                  & 0.861 $\pm$ 0.055 & 0.837 $\pm$ 0.060 & 0.873 $\pm$ 0.055 & 0.706 $\pm$ 0.075 \\ \hline
\textbf{Avg. Score}              & 0.917 $\pm$ 0.057 & 0.879 $\pm$ 0.055 & 0.901 $\pm$ 0.059 & 0.667 $\pm$ 0.104 \\ \Xhline{1pt}
\end{tabular}
\label{table:intra}
\end{table}

\begin{table}[t!]
\renewcommand{\arraystretch}{1.6}
\caption{Intersession test accuracy across Experiment 2 settings and feature sets (and TVGGNet), averaged across subjects in DB$_{b}$.}
\footnotesize
\setlength{\tabcolsep}{6pt}
\begin{tabular}{ccccc}
\Xhline{1pt}
                        & AVS                 & CS                 & AC                 \\ \hline
TD                      & 0.407 $\pm$0.199  & 0.343 $\pm$ 0.142 & 0.198 $\pm$ 0.108 \\ 
TD (PCA)                & \textbf{0.416 $\pm$0.198}  & 0.393 $\pm$ 0.132 & 0.368 $\pm$ 0.219 \\ 
\hline
ETD                     & 0.425 $\pm$ 0.211 & 0.363 $\pm$ 0.162 & 0.226 $\pm$ 0.115 \\ 
ETD (PCA)               & \textbf{0.437 $\pm$ 0.209} & 0.367 $\pm$ 0.153 & 0.414 $\pm$ 0.215 \\ 
\hline
NinaPro                 & 0.407 $\pm$ 0.205 & 0.335 $\pm$ 0.158 & 0.205 $\pm$ 0.105 \\ 
NinaPro (PCA)           & \textbf{0.425 $\pm$ 0.170} & 0.394 $\pm$ 0.125 & 0.396 $\pm$ 0.216 \\ 
\hline
SampEn                  & \textbf{0.448 $\pm$ 0.217} & 0.369 $\pm$ 0.173 & 0.229 $\pm$ 0.111 \\ 
SampEn (PCA)            & 0.412 $\pm$ 0.204 & 0.385 $\pm$ 0.163 & 0.421 $\pm$ 0.220 \\ 
\hline
TVGGNet                 & \textbf{0.420 $\pm$ 0.183} & 0.365 $\pm$ 0.156 & 0.394 $\pm$ 0.192 \\ 

\hline
\textbf{Avg. Score}        & 0.421 $\pm$ 0.198 & 0.355 $\pm$ 0.155 & 0.250  $\pm$ 0.146 \\
\textbf{Avg. Score (PCA)}        & 0.422 $\pm$ 0.188 & 0.381 $\pm$ 0.143 & 0.399 $\pm$ 0.207 \\ \Xhline{1pt}
\end{tabular}
\label{table:inter}
\end{table}

\subsection{Intersession Performance}
Given that electrode shifts can result in significant performance degradation across sessions \cite{young2011effects}, the effect of our electrode shift generalization method was evaluated on intersession performance. For each subject, data from one session was added to the training set and data from the remaining session was added to the testing set. Three conditions were considered: 
\begin{enumerate}
    \item \textbf{(AVS)}: training on all valid channel subsets and testing on central channel subset
    \item \textbf{(CS)}: training/testing on central channel subset
    \item \textbf{(AC)}: training and testing with a model that used all (128) channels as input, as a naive method to levarage HD-sEMG
\end{enumerate}


Given the high input-dimensionality of condition \textbf{(AC)}, performance was also evaluated with principal components analysis (PCA) dimensionality reduction by keeping principal components as to retain 95\% of explained variance across conditions. As shown in Table \ref{table:inter}, the use PCA resulted in higher test performance for all feature sets included and conditions, with the exception of the SampEn feature set under condition \textbf{(AVS)}, as its low feature count was sufficiently regularized by training on all valid channels subsets. When considering these feature sets without PCA, given their test accuracies are not significantly different ($F(3, 212)=0.441, p=0.724$), the average score across feature sets was considered. 

Paired t-tests showed significantly positive mean differences of $+6.9\%$ between \textbf{(AVS)} and \textbf{(CS)} ($t(17)=3.932, p<0.001$), and of $+20.7\%$ between \textbf{(AVS)} and \textbf{(AC)} ($t(17)=5.438, p<0.001$). This suggests that without dimensionality reduction, given the overfitting tendency of the large input dimensionality present, training an 8 input channel model with data from all valid subsets results in significantly greater generalization. This is especially the case when comparing with condition \textbf{(AC)}, where performance is far lower than in other cases. The following analysis only includes results using PCA, with the exception of TVGGNet, as it already learns its own feature space remapping. 

As before, given a one-way ANOVA test suggested that the implemented feature sets and the TVGGNet do not exhibit significantly different test accuracies ($F(3,284)=1.640, p=0.164$), the average score across feature sets and TVGGNET was considered in this analysis. While there was a significantly positive mean difference of $+3.7\%$ between \textbf{(AVS)} and \textbf{(CS)} ($t(17)=1.929,p=0.035$), there was no significant mean difference between \textbf{(AVS)} and \textbf{(AC)} ($t(17)=0.966, p=0.174$). However, despite the lack of significance, \textbf{(AVS)} maintained performance with 16x less input channels. Furthermore, there was an average improvement of $2.2\%$ even when excluding the use of PCA in condition \textbf{(AVS)}. Therefore, training across all valid channel subsets results in significant performance improvements even without PCA, and performs comparably or better than other conditions with far lower computational complexity.

\section{CONCLUSION}
\label{sec:con}

We have investigated a novel use of HD-sEMG for gesture recognition where only 8 channels were used as inputs, and the remaining recorded channels were used for augmenting the training dataset, as opposed to working with high-dimensional inputs. This method allows different feature sets and networks to generalize across different electrode placements, reducing overfitting and consequently leading to significant improvements in intersession performance relative to other HD-sEMG-based models without the need of dimensionality reduction. Given the high variability in intersession performance improvement across subjects, it is likely that in some cases, other confounding factors such as gesture technique, limb position and contraction level, are the largest contributors to performance degradation, occluding the true robustness benefits of the proposed method.

One limitation of this study is that only proximal-distal shifts were considered, whereas clockwise/anticlockwise shifts typically result in larger performance drops \cite{electrode-shift}. Preliminary experiments showed that training on these shifts did not contribute to generalization, likely due to shifts across the muscle having a larger effect on EMG signal properties than along the muscle. Thus, future studies should use larger HD-sEMG datasets to train larger, more expressive networks to evaluate the generalization across these shifts. Furthermore, beyond robustness and computational complexity benefits, the input dimensionality chosen will enable future research on domain adaptation across acquisition systems with publicly available datasets, such as the ones within the Ninapro database \cite{ninapro}. This may open a new avenue for HD-sEMG in gesture recognition: acquiring complex training datasets that can be leveraged to develop robust algorithms for simpler, inexpensive and commercially viable systems.

\bibliographystyle{IEEEbib}
{\small
\bibliography{strings,refs}}

\end{document}